\documentclass[10pt,twocolumn,letterpaper]{article}

\usepackage{cvpr}              
\usepackage{multirow}
\usepackage{makecell}
\usepackage{pifont}
\usepackage{arydshln}
\newcommand{\cmark}{\ding{51}}%
\newcommand{\xmark}{\ding{55}}%
\usepackage{booktabs}
\usepackage{balance}
\usepackage{subcaption}
\usepackage{graphicx}
\usepackage{marvosym}

\definecolor{cvprblue}{rgb}{0.21,0.49,0.74}
\usepackage[pagebackref,breaklinks,colorlinks,allcolors=cvprblue]{hyperref}

\def\paperID{11081} 
\def\confName{CVPR}
\def\confYear{2025}

\title{\texttt{Neuron}: Learning Context-Aware Evolving Representations for Zero-Shot Skeleton Action Recognition}

\author{Yang Chen\textsuperscript{1}, Jingcai Guo\textsuperscript{1\thanks{Jingcai Guo is the corresponding author.}}, Song Guo\textsuperscript{2}, Dacheng Tao\textsuperscript{3}\\
\textsuperscript{1}Department of Computing, The Hong Kong Polytechnic University, Hong Kong SAR\\
\textsuperscript{2}Department of CES, The Hong Kong University of Science and Technology, Hong Kong SAR\\
\textsuperscript{3}College of Computing and Data Science, Nanyang Technological University, Singapore\\
{\Letter:~\tt jc-jingcai.guo@polyu.edu.hk}
}

\begin{document}
\maketitle
\begin{abstract}
Zero-shot skeleton action recognition is a non-trivial task that requires robust unseen generalization with prior knowledge from only seen classes and shared semantics. 
Existing methods typically build the skeleton-semantics interactions by uncontrollable mappings and conspicuous representations, thereby can hardly capture the intricate and fine-grained relationship for effective cross-modal transferability. 
To address these issues, we propose a novel dy\underline{\textbf{N}}amically \underline{\textbf{E}}volving d\underline{\textbf{U}}al skeleton-semantic syne\underline{\textbf{R}}gistic framework with the guidance of c\underline{\textbf{O}}ntext-aware side informatio\underline{\textbf{N}} (dubbed \textbf{Neuron}), to explore more fine-grained cross-modal correspondence from micro to macro perspectives at both spatial and temporal levels, respectively. 
Concretely, 1) we first construct the spatial-temporal evolving micro-prototypes and integrate dynamic context-aware side information to capture the intricate and synergistic skeleton-semantic correlations step-by-step, progressively refining cross-model alignment; and 2) we introduce the spatial compression and temporal memory mechanisms to guide the growth of spatial-temporal micro-prototypes, enabling them to absorb structure-related spatial representations and regularity-dependent temporal patterns. 
Notably, such processes are analogous to the learning and growth of neurons, equipping the framework with the capacity to generalize to novel unseen action categories. 
%
Extensive experiments on various benchmark datasets demonstrated the superiority of the proposed method~\footnote{The code is available at: \url{https://anonymous.4open.science/r/Neuron}.}.
\end{abstract}    
\vspace{-10pt}
\section{Introduction}
\label{sec:intro}

\begin{figure}[!t]
  \centering
  \includegraphics[width=0.88\linewidth]{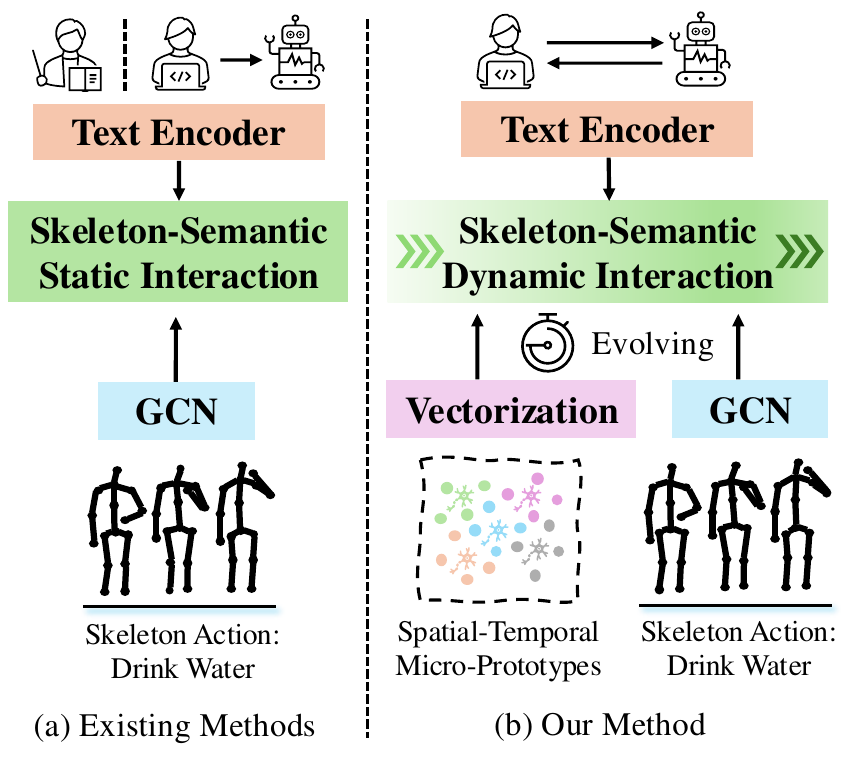}
  \vspace{-5pt}
  \caption{Method sketches: (a) Existing methods employ homogeneous semantics (manual-designed or one-turn LLMs) to align the skeleton space with one-step (\textcolor{blue}{static}); (b) Our Neuron introduces the context-aware side information (multi-turn LLMs) and evolving micro-prototypes to capture cross-modal correspondence from micro to macro perspectives (\textcolor{red}{dynamic}) for controllable alignment.}
  \label{fig:overview}
  \vspace{-15pt}
\end{figure}

Action recognition has gained considerable attention for years due to its broad applications, \textit{e.g.}, sports analysis \cite{Xu_2024_CVPR}, rehabilitation assessment \cite{he2024expert}, anomaly action recognition \cite{sato2023prompt}, etc. Benefiting from the excellent privacy-preserving and data-efficiency properties of skeleton modality, a wealth of innovative research has sprung up to advance this field. Among these developments, several challenging and realistic issues have emerged, notably spurring interest in zero-shot skeleton action recognition~\cite{chen2024fine,li2024sadvae, zhu2024part}.

Conventional supervised and self-supervised skeleton-based action recognition methods \cite{chen2024fine, chen2024c2vl} are usually hindered by their dependence on expensive annotated data and limited generalization to novel unseen categories. These methods generally confine their classification abilities to actions present in both training and test sets \cite{zhou2024blockgcn, zhou2023learning, chen2021channel, cheng2020skeleton}.
In contrast, zero-shot skeleton action recognition effectively addresses the above dilemmas by leveraging only limited seen skeleton samples and a predefined semantic corpus, thus enabling the recognition of previously unseen action categories \cite{chen2024fine, li2024sadvae, zhu2024part, zhou2023zero, li2023multi, gupta2021syntactically, jasani2019skeleton}.
Furthermore, such a zero-shot learning (ZSL) setting can be extended to generalized zero-shot learning (GZSL), wherein both seen and unseen categories are recognized, making the task more challenging yet more applicable to real-world scenarios. 
The common solutions involve establishing robust cross-modal relationships by embedding skeleton and semantic data within a joint space, forming a bridge that transfers knowledge from seen to unseen actions \cite{jasani2019skeleton, gupta2021syntactically, li2023multi, zhou2023zero}. In pursuit of improved generalization, recent studies have further sought ways for fine-grained alignment by decomposing representations either explicitly \cite{chen2024fine, zhu2024part} or implicitly \cite{li2024sadvae}.

Despite the progress made, crucial challenges remain unsolved, \textit{i.e.}, 
1) existing semantics are limited to manual-designed category names or one-turn LLMs-generated action descriptions (Fig.~\ref{fig:overview}~(a)), leading to high homogeneity semantics among similar actions. Taking ``\textit{kicking}" and ``\textit{side kick}" as samples, homogeneous semantics have made the joint embedding space indivisible among these highly similar inter-class actions. Additionally, simple action descriptions cannot encapsulate the intra-class diversity of skeletons, where spatial variations (\textit{e.g.}, movement scale) and temporal differences (\textit{e.g.}, speed) vary across individuals; 
and 2) directly aligning skeleton and semantic features into a joint embedding space lacks fine-grained control over this process, which causes the model to overemphasize irrelevant or shortcut semantics, \textit{i.e.}, they cannot locate the vital skeleton features with intended semantics correctly, resulting in suboptimal cross-modal fine-grained correspondence. 
Inspired by the growth process of neurons from small to large that can learn and adapt incrementally through repeated stimuli to form complex connections that enable generalization, we consider designing a dynamic and controllable fine-grained aligning framework from micro to macro perspectives that can refine the interactions incrementally by the guidance of context-aware side information, 
allowing better generalization to unseen actions.

Specifically, we propose a novel dynamically evolving dual skeleton-semantic synergistic framework via the guidance from context-aware side information, termed \textit{\textbf{Neuron}}, to enable controlled fine-grained cross-modal interactions (Fig.~\ref{fig:overview}~(b)). Firstly, we employ LLMs (\textit{e.g.}, GPT-4o) with multi-turn interactive prompts to generate enriched, contextual side information, providing detailed semantic descriptions of skeletons that span spatial and temporal variations. Spatial variations range from coarse to mid- and fine-grained structural granularity, while temporal dynamics cover the evolution of movements from start to mid and end phases. With this contextual semantic guidance, we construct spatial-temporal micro-prototypes to capture the synergistic skeleton-semantic relationship iteratively. The spatial micro-prototype evolves through the spatial compression mechanism, guided by stepwise semantic granularity upgrades to isolate discriminative structure-related spatial representations. Similarly, the temporal micro-prototype grows dynamically through the temporal memory mechanism, directed by temporal phase-evolution semantics in turn, to search effective regularity-dependent temporal patterns. Both micro-prototypes are updated iteratively from the micro to the macro perspectives, emulating neuron-like incremental growth to refine their understanding of skeleton actions.
This process enhances cross-modal transferability, empowering the framework to generalize effectively to unseen action categories.

The main contributions can be summarized as follows:
\begin{itemize}[leftmargin=2em]
    \item We introduce dynamic, context-aware side information to guide the spatial-temporal micro-prototypes in evolving from micro to macro perspectives, achieving controlled, fine-grained cross-modal alignment.
    \item We propose the spatial compression and temporal memory mechanisms, enabling the micro-prototypes to explore structure-related spatial representations and regularity-dependent temporal patterns.
    \item Extensive experiments show that the proposed method achieves state-of-the-art results in both ZSL and GZSL settings on NTU RGB+D, NTU RGB+D 120, and PKU-MMD datasets.
\end{itemize}
\section{Related Work}
\label{sec:relatedwork}

\begin{table}
    \begin{center}
    \resizebox{.47\textwidth}{!}{
        \begin{tabular}{cc|cc|ccc|c}
        \hline
        \multicolumn{2}{c|}{\textbf{Alignment}} & \multicolumn{2}{c|}{\textbf{Adaptability}} & \multicolumn{3}{c|}{\textbf{Semantic Source}} & \multirow{2}{*}{\textbf{Method}}\\
        GL & FG & ST & DY & {\color{blue}\textbf{Manual}} & {\color{green}\textbf{H-LLMs}}& {\color{red}\textbf{C-LLMs}}&\\
        \hline \hline
        \cmark & \xmark & \cmark & \xmark & \cmark & \xmark & \xmark & RelationNet \cite{jasani2019skeleton}, SynSE \cite{gupta2021syntactically}\\
        \cmark & \xmark & \cmark & \xmark & \cmark & \cmark & \xmark & SMIE \cite{zhou2023zero}\\
        \xmark & \cmark & \cmark & \xmark & \cmark & \cmark & \xmark & PURLS \cite{zhu2024part}, STAR \cite{chen2024fine}\\
        \xmark & \cmark & \cmark & \xmark & \cmark & \xmark & \xmark & SA-DAVE \cite{li2024sadvae}\\
        \xmark & \cmark & \xmark & \cmark & \xmark & \xmark & \cmark & \textbf{Neruon (ours)}\\
        \hline
        \end{tabular}}
    \end{center}
    \vspace{-10pt}
    \caption{Comparison with existing methods. The GL and FG are denoted as global and fine-grained. The ST and DY are represented as static and dynamic. {\color{blue}Manual}, {\color{green}H-LLMs} and {\color{red}C-LLMs} represent manual-designed category names, LLMs-generated homogeneous descriptions, and context-aware side information, indicating the relatively {\color{blue}Low}, {\color{green}Mid} and {\color{red}High} semantic strength.}
    \label{tab:zsar}
    \vspace{-10pt}
\end{table}

\begin{figure*}[t!]
  \centering
  \includegraphics[width=\linewidth, height=0.35\linewidth]{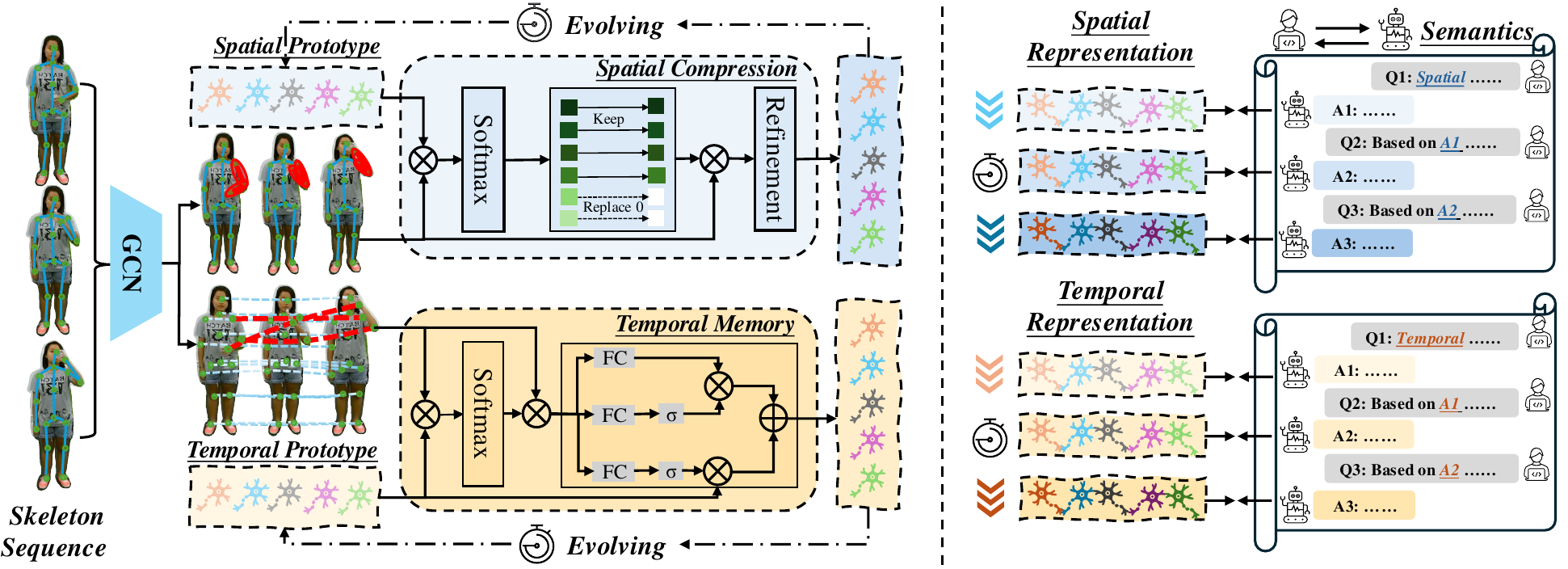}\\
  \hspace{70pt}\small{(a) Evolving Representation Learning} \hspace{100pt} \small{(b) Stepwise Cross-Modal Alignment} \\
  \vspace{-5pt}
  \caption{The pipeline of the proposed method. (a) represents the evolving spatial-temporal representation learning process. (b) shows the stepwise skeleton-semantic alignment.}
  \vspace{-10pt}
  \label{fig.framework}
\end{figure*}

\subsection{Zero-Shot Skeleton Action Recognition}
Zero-shot skeleton action recognition methods aim to establish the relationship between the seen skeletons and their associated semantics to recognize novel action categories.
A pioneering approach, RelationNet \cite{jasani2019skeleton}, identified novel categories by learning deep non-linear metrics between global skeleton features and category names in an embedding manner. Differently, SynSE \cite{gupta2021syntactically} paves the generative way using VAEs to align global skeleton features with verb-noun embeddings of category names. SMIE \cite{zhou2023zero} employs the temporal constraint with maximizing mutual information to achieve global alignment. However, these methods rely on global alignment with manual-designed category names, limiting the granularity and diversity of captured semantics. 
Recently, PURLS \cite{zhu2024part} and STAR \cite{chen2024fine} explicitly decompose skeletons into overlapping parts with LLMs-generated part-specific descriptions, enabling fine-grained alignment. 
Unfortunately, these generated descriptions are homogenous across actions, reducing their discriminative capacity. SA-DAVE \cite{li2024sadvae} improves generalization by decoupling skeleton representations into semantic-related and irrelevant components. 
\textbf{[\textit{Summary}]}: Unlike prior methods, our proposed framework advances zero-shot skeleton action recognition in two aspects. First, we utilize the context-aware side information generated through multi-turn interactive prompting on LLMs, providing more discriminative, diverse, and context-rich semantics that go beyond manual-designed category names or homogeneous descriptions generated by one-turn interaction. Second and more importantly, we design a dynamic, evolving, and synergistic alignment framework to explore better cross-modal correspondence, contrasting with prior methods that employ static, fixed, and isolated alignment processes, as shown in Table \ref{tab:zsar}.

\vspace{-3pt}

\subsection{Multi-modal Skeleton Representation Learning}
Inspired by the success of CLIP \cite{radford2021learning}, numerous studies have explored multi-modal fusion or cross-model distillation strategies to enhance skeleton representation capacity. These approaches incorporate data from complementary modalities, such as RGB videos \cite{liu2024multi, song2020modality, bruce2022mmnet, kim2023cross, chen2024c2vl}, depth sequences \cite{cui2023multi}, text descriptions \cite{lu2024cross, chen2024c2vl, xiang2023generative}, flow information \cite{sun2023unified,mao2022cmd}, and sensor signals \cite{ni2022progressive}. The core of the multi-modal fusion methods \cite{he2024enhancing, liu2024multi,song2020modality, sun2023unified,cui2023multi,bruce2022mmnet,kim2023cross} is to compensate for the shortcomings of skeletons by relying on the strengths of other modalities, while the cross-modal distillation methods \cite{chen2024c2vl,xiang2023generative,lu2024cross,mao2022cmd,ni2022progressive} aim to capture modality-agnostic representations for action recognition. Nevertheless, both methods lack explicit control over the representation learning process, resulting in models that capture shortcut features. Recently, some methods have made strides in controlled representation learning \cite{bleeker2022multi,dancette2021beyond,lu2024enhancing}, yet not in the skeleton action recognition area. \textbf{[\textit{Summary}]}: In this paper, we propose the spatial compression and temporal memory mechanisms that guide the constructed micro-prototypes to capture skeleton representations in a controlled manner, progressing from micro to macro perspectives. This approach allows the model to focus on intended semantics and avoid being stuck into shortcut features, thereby enhancing the transferability and robustness of learned representations.


\section{Method}

\subsection{Problem Formulation}
For the zero-shot skeleton action recognition problem, the skeleton dataset $\mathcal{D}=\{x_{i}, y_{i}, a_{i}\}_{i=1}^{N}$ with $\left | \mathcal{Y}  \right |$ categories can be divided into three overlapping parts: the train-set $\mathcal{D}_{tr}^{s}$ with seen categories $\mathcal{Y}^{s}$, the test-set $\mathcal{D}_{te}^{u}$ with unseen categories $\mathcal{Y}^{u}$, and another test-set $\mathcal{D}_{te}^{s}$ with seen categories $\mathcal{Y}^{s}$. Here, $x_{i}\in \mathbb{R}^{3\times T\times V\times M }$ represents the skeleton sequence, where 3 corresponds to the 3D coordinates of human joints, $T$ denotes the sequence frames, $V$ is the number of human joints, and $M$ indicates the human numbers. The symbol $y_{i} \in \mathcal{Y}$ is the corresponding action label, where $a_{i} \in \mathcal{A}$ signifies the associated semantics. Notably, $\mathcal{Y} = \mathcal{Y}^{s} \cup \mathcal{Y}^{u}$ and $\mathcal{Y}^{s} \cap \mathcal{Y}^{u}=\varnothing$. For each action category, a manual-designed category name and one-turn LLMs-generated action description are adopted as the semantics $a_{i}$ in prior works \cite{jasani2019skeleton, gupta2021syntactically, zhou2023zero, zhu2024part, chen2024fine, li2024sadvae}. During the training phase, we establish the relationship between the $x_{i}^{s}$ and $a_{i}^{s}$ within the train-set $\mathcal{D}_{tr}^{s}$. In the inference phase, we need to predict categories in both $\mathcal{D}_{te}^{u}$ and $\mathcal{D}_{te}^{s} \cup \mathcal{D}_{te}^{u}$ under the ZSL and GZSL settings through the above-constructed relationship, respectively. For simplicity, we omit the superscript for seen (s) and unseen (u) categories in the following sections.

\subsection{Contextual Side Information Generation}
We generate more heterogeneous and multifaceted contextual semantic descriptions of skeletons from the perspective of spatial structure granularity and temporal resolution variation than previous methods \cite{jasani2019skeleton, gupta2021syntactically,li2024sadvae, zhou2023zero, zhu2024part, chen2024fine}. 
Specifically, we engage the LLM (e.g., GPT-4o) to produce semantic descriptions of skeleton movements in multi-turn inquiries, including the hierarchical spatial structure and evolving temporal resolutions, as shown in Fig. \ref{fig.framework} (b). 
In the $e$-th phase, we can obtain $N_a$ state descriptions $\mathcal{A}_{y}^{e}$ for the same skeleton category $y$, represented by different individuals, thereby enriching the intra-class diversity of semantics. With these step-by-step queries, the semantics among the categories become more discriminative, prompting inter-class separability of semantics. Upon this iterative process, we enhance the coherent and contextual representation capabilities of the semantics $\mathcal{A}=\{ \mathcal{A}_{y}^{1},...,\mathcal{A}_{y}^{e},...,\mathcal{A}_{y}^{N_e}  \}_{y=1}^{|\mathcal{Y}|}$, where $N_e$ is the number of phase. Meanwhile, the evolving property of semantics from LLMs can guide the learning process of skeleton representation for better cross-modal correspondences. Afterward, we utilize pre-trained text encoder $\Psi (\cdot )$ to extract the contextual semantic features $\mathcal{Z}=\{ \mathcal{Z}_{y} \}_{y=1}^{|\mathcal{Y}|}$, where $\mathcal{Z}_{y}=\{\mathcal{Z}_{y}^{s_e}, \mathcal{Z}_{y}^{t_e} \in \mathbb{R}^{N_a \times d} \}_{e=1}^{N_e}$, where $\mathcal{Z}_{y}^{s_e}$ and $\mathcal{Z}_{y}^{t_e}$ are the spatial and temporal semantics of $y$-th action category at $e$-th phase, d is the dimension of semantic feature. In this work, we set the $N_e=3$.

\subsection{Spatial-Temporal Synergistic Growing}
\vspace{0.5em}
\noindent\textbf{Spatial-Temporal Micro-prototypes Construction.}
Decoupling spatial structures and temporal variations is vital in distinguishing similar actions. 
As described in \cite{zhou2023learning}, temporal variations help differentiate "put things into a bag" and "take things out of a bag", while spatial structures are significant for distinguishing them from "reach into pockets". 
Thus, we separate spatial and temporal representations for alignment independently. First, we extract the skeleton representations $\mathcal{F}=\Phi (x) \in \mathbb{R}^{\widehat{T}\times \widehat{V}\times C}$ by skeleton encoder $\Phi (\cdot )$, where $\widehat{T}\times \widehat{V}$ represents the spatial-temporal dimension after transformations, $C$ is the feature dimension. Then, we pool the temporal and spatial dimensions respectively to obtain the spatial feature $\mathcal{F}_{s} \in \mathbb{R}^{\widehat{V}\times C}$ and the temporal feature $\mathcal{F}_{t} \in \mathbb{R}^{\widehat{T}\times C}$. To effectively align the skeleton and semantic spaces, we design the learnable spatial micro-prototype $\mathcal{P}_{s}=\{p_{k}^{s} \in \mathbb{R}^{C \times 1} \}_{k=1}^{N_s}$ to capture the patterns of skeleton topology structure and construct the learnable temporal micro-prototype $\mathcal{P}_{t}=\{p_{k}^{t} \in \mathbb{R}^{C \times 1} \}_{k=1}^{N_t}$ to record the temporal motion variations of the skeleton, where $p_{k}^{s}$ and $p_{k}^{t}$ are randomly initialized attributes of skeleton representation (e.g., amplitude and velocity \cite{chen2024fine}), $N_s$ and $N_t$ are the number of primitives. Guided by semantics $\mathcal{Z}$, $\mathcal{P}_{s}$ and $\mathcal{P}_{t}$ progressively learn useful skeleton representations, similar to the growth process of neurons.

\vspace{0.5em}
\noindent\textbf{Spatial Compression Mechanism.}
For the given spatial micro-prototype $\mathcal{P}_s$, we aim to enable it to progressively and controllably absorb structure-related spatial patterns from skeletons across three phases, enhancing the cross-modal correspondence at the spatial level from coarse-grained to fine-grained. Specifically, in the first phase, we treat the initialized $\mathcal{P}_{s}^{0}$ as the query matrix to calculate similarity scores with spatial skeleton representations $\mathcal{F}^s$:
\begin{equation}
    \mathcal{H}^{s_0}=softmax(\mathcal{F}_s \mathcal{P}_s^0),
\end{equation}
where $h_{j,k}^{s_0} \in \mathcal{H}^{s_0}$ denotes the similarity between the $j$-th joint feature and the $k$-th spatial micro-prototype attribute at the coarse-grained level. Normally, a straightforward approach is to aggregate the joint features to obtain the updated micro-prototype for alignment with semantics:
\begin{equation}
    \mathcal{P}_s^1 = \varphi_s  (\mathcal{F}_s^{T}\mathcal{H}^{s_0}),
    \label{eq:aggerate}
\end{equation}
where $\varphi_s(\cdot)$ is the refinement function composed of MLPs. However, not all the micro-prototype attributes are relevant for each joint across different samples; aggregating all attributes may inadvertently overemphasize those unrelated to the action itself (e.g., camera views). To mitigate this issue, we selectively drop joint-unrelated attributes, thus avoiding shortcut feature learning. We sort the matrix $ \mathcal{H}^{s_0}$ row-wise and partition it into two subsets with the hyper-parameter $\alpha \in [0, 1]$. Then, we maintain the top-$\alpha$ scores and replace others with zeros. Using the modified matrix $\mathcal{\hat{H}}^{s_0}$ in Eq \ref{eq:aggerate}, we can get the updated micro-prototype $\mathcal{P}_s^1$. Considering coarse-grained exploration alone is insufficient, we repeat the above operations at the mid-grained and fine-grained levels further. Meanwhile, $\alpha$ is incrementally increased in each phase to allow for more precise attribute selection. Through these controllable compression processes, the spatial micro-prototype is growing to $\mathcal{P}_s^3$, capturing more general and concise spatial pattern representations.

\vspace{0.5em}
\noindent\textbf{Temporal Memory Mechanism.}
Compared to the spatial micro-prototype $\mathcal{P}_{s}$ growing through the refinement of hierarchical structure, the temporal micro-prototype $\mathcal{P}_{t}$ develops alongside the evolution of resolution. Unlike the complete spatial feature $\mathcal{F}_s$, we split the temporal feature into three phases to obtain segment features $\mathcal{F}_t=\{\mathcal{F}_t^{e} \in \mathbb{R}^{\frac{\hat{T}}{3} \times C}\}_{e=1}^{N_e}$ for progressive alignment. Initially, we calculate the similarity scores between the initialized $P_t^0$ and 1-st temporal feature $F_t^1$, and then aggregate all frame features to update the temporal micro-prototype as follows:
\begin{align}
    \mathcal{H}^{t_0}&=softmax(\mathcal{F}_t^{0} \mathcal{P}_t^0),\\
    \mathcal{\hat{P}}_t^0 &= (\mathcal{F}_t^{0})^{T}\mathcal{H}^{t_0},
\end{align}
We do not adopt the compression mechanism here because the sequential temporal features in each phase can constrain the learning process not stuck into the short features. However, the updated temporal micro-prototype in the $e$-th phase faces the challenge of knowledge oblivion that is learned in the $(e-1)$-th phase, as shown in the first row of Fig. \ref{fig:memory} (a) - \ref{fig:memory} (c). To address this, in each phase, we introduce the memory mechanism to remember new information while selectively recalling the knowledge from the previous phase:
\begin{equation}
    \mathcal{P}_t^1 = \underbrace{\sigma(\rho_r(\mathcal{\hat{P}}_t^0)) \cdot \mathcal{P}_t^0}_{recall} + \underbrace{\sigma(\rho_m(\mathcal{\hat{P}}_t^0)) \cdot \varphi_t(\mathcal{\hat{P}}_t^0)}_{remember},
\end{equation}
where $\sigma(\cdot)$ is the sigmoid function to quantize the remember and recall probability, $\rho_r(\cdot)$ and $\rho_m(\cdot)$ are MLPs for projection, $\varphi_t(\cdot)$ is the refine functions. Through the evolution from the first to the third phase, the micro-prototype $\mathcal{P}_t^0$ progressively captures the regularity-dependent temporal clues, controllable forming to the $\mathcal{P}_t^3$ at the macro level.

\subsection{Overall Objective}
To align the skeleton and semantic spaces at the fine-grained level, we start from the coarse-grained level, step-by-step, instructing micro-prototypes to update iteratively and controllably from micro to macro perspectives. For each $e$-th phase, the spatial-temporal micro-prototypes $\mathcal{P}_s^e \in \mathbb{R}^{C \times N_s}$ and $\mathcal{P}_t^e \in \mathbb{R}^{C \times N_t}$ contain the specific skeleton patterns for different attributes, so we pool all the channels to produce the skeleton representation $\mathcal{X}_s^{e},  \mathcal{X}_t^{e} \in \mathbb{R}^{C}$. Meanwhile, we pool all the contextual semantic features $\mathcal{Z}_y^{s_e}, \mathcal{Z}_y^{t_e} \in \mathbb{R}^{Na \times d }$ into $\mathcal{\hat{Z}}_y^{s_e}, \mathcal{\hat{Z}}_y^{t_e} \in \mathbb{R}^d$. Then, we define the alignment operation as:
\begin{align}
    \mathcal{L}_s^e = -log(\frac{exp(\phi_s(\mathcal{X}_s^e)^T \psi(\mathcal{\hat{Z}}_y^{s_e}))}{\sum_{k \in \mathcal{Y}^s  }exp(\phi_s(\mathcal{X}_s^e)^T \psi(\mathcal{\hat{Z}}_k^{s_e}))} ), \\
    \mathcal{L}_t^e = -log(\frac{exp(\phi_t(\mathcal{X}_t^e)^T \psi(\mathcal{\hat{Z}}_y^{t_e}))}{\sum_{k \in \mathcal{Y}^s  }exp(\phi_t(\mathcal{X}_t^e)^T \psi(\mathcal{\hat{Z}}_k^{t_e}))} ),
\end{align}
where $\phi_s(\cdot)$, $\phi_t(\cdot)$, and $\psi(\cdot)$ are project functions. Last, the overall optimization objective is defined as:
\begin{equation}
    \mathcal{L} = \sum_{e=1}^{N_e}(\mathcal{L}_s^e + \mathcal{L}_t^e).
\end{equation}

\subsection{ZSL/GZSL Prediction}
In the inference stage, we can contain the spatial-temporal features $\mathcal{X}_s^{m},  \mathcal{X}_t^{m}$ that are updated by micro-prototypes. Then, we can predict the unseen skeleton category by calculating the similarity with unseen semantics. Furthermore, the calibrated stacking method \cite{chao2016empirical} is employed to mitigate the domain shift for GZSL prediction. The whole process is formulated as follows:
\begin{align}
    \hat{y}^{s}=\mathop{arg\max}\limits_{y\in\mathcal{Y}^{u}/\mathcal{Y}}^{} \rho_s(\mathcal{X}_s^3)^T \phi(\mathcal{\hat{Z}}_y^{s_3})-\gamma_s\mathbb{I}[y\in \mathcal{Y}^{s}],\\
    \hat{y}^{t}=\mathop{arg\max}\limits_{y\in\mathcal{Y}^{u}/\mathcal{Y}}^{} \rho_t(\mathcal{X}_t^3)^T \phi(\mathcal{\hat{Z}}_y^{t_3})-\gamma_t\mathbb{I}[y\in \mathcal{Y}^{s}],
\end{align}
where $\mathcal{Y}^{u}/\mathcal{Y}$ corresponds to the ZSL/GZSL setting, and $\gamma_s/\gamma_t$ are calibration factors. As mentioned above, different actions have different propensities for spatial and temporal, so we define the final prediction as $\hat{y}^* = \{\hat{y}^{s}, \hat{y}^{t} \}$.

\section{Experiments}
\subsection{Datasets}

\vspace{0.5em}
\noindent\textbf{NTU RGB+D 60} \cite{shahroudy2016ntu}.
It contains 56880 skeleton sequences with 60 action categories performed by 40 subjects captured from 3 distinct views. Generally, two benchmarks are commonly utilized for evaluation: cross-subject (Xsub) and cross-view (Xview). The Xsub splits all sequences based on subject indices, with 20 subjects allocated for training and the remaining 20 subjects for testing. The Xview divides data based on view variations, using views 2 and 3 for training while reserving view 1 for testing.

\vspace{0.5em}
\noindent\textbf{NTU RGB+D 120} \cite{liu2019ntu}.
As an extended version of the NTU RGB+D 60 dataset, it encompasses 114,480 skeleton sequences with 120 action categories. Similarly, it also provides the two official benchmarks: cross-subject (Xsub) and cross-setup (Xset). The Xsub task requires 53 subjects for training, with the remaining subjects designated for testing. Meanwhile, the Xset task utilizes data captured from even camera IDs for training and others for testing.

\vspace{0.5em}
\noindent\textbf{PKU-MMD} \cite{liu2017pku}.
Almost 20000 skeleton sequences with 51 categories are collected in this dataset, which can be split into two phases with increasing challenges. It offers two benchmarks: cross-subject (Xsub) and cross-view (Xview). For the Xsub task, 57 subjects are designated for training, while 9 subjects are reserved for testing. For Xview, the middle and right views data is included in the training set, with the left view used for testing. In this work, we utilize the first phase for experiments following the \cite{zhou2023zero, chen2024fine}.

\begin{table*}
  \caption{State-of-the-art comparisons on NTU RGB+D 60 dataset under the ZSL and GZSL setting. The best and the second-best results are marked in \textcolor{red}{\textbf{Red}} and \textcolor{blue}{\textbf{Blue}}, respectively.}
  \vspace{-10pt}
  \label{table:ntu60}
  \centering
\resizebox{1.0\textwidth}{!}{
  \begin{tabular}{l|c|ccc|c|ccc|c|ccc|c|ccc}
    \hline
    \multirow{4}{*}{Method} & \multicolumn{8}{c|}{Xsub} & \multicolumn{8}{c}{Xview}\\
    \cline{2-17} 
     & \multicolumn{4}{c|}{55/5 Split} & \multicolumn{4}{c|}{48/12 Split} & \multicolumn{4}{c|}{55/5 Split} & \multicolumn{4}{c}{48/12 Split}\\
    \cline{2-17}
    & ZSL & \multicolumn{3}{c|}{GZSL} &  ZSL & \multicolumn{3}{c|}{GZSL} &  ZSL & \multicolumn{3}{c|}{GZSL} &  ZSL & \multicolumn{3}{c}{GZSL} \\
    \cline{2-17}
     & $\mathcal{A}cc$ & $\mathcal{S}$ & $\mathcal{U}$ & $\mathcal{H}$ & $\mathcal{A}cc$ & $\mathcal{S}$ & $\mathcal{U}$ & $\mathcal{H}$ & $\mathcal{A}cc$ & $\mathcal{S}$ & $\mathcal{U}$ & $\mathcal{H}$ & $\mathcal{A}cc$ & $\mathcal{S}$ & $\mathcal{U}$ & $\mathcal{H}$ \\
    \hline
    ReViSE \cite{hubert2017learning} & 69.5 & 40.8 & 50.2 & 45.0 & 24.0 & 21.8 & 14.8 & 17.6 & 54.4 & 25.8 & 29.3 & 27.4 & 17.2 & 34.2 & 16.4 & 22.1 \\
    JPoSE \cite{wray2019fine} & 73.7 & 66.5 & 53.5 & 59.3 & 27.5 & 28.6 & 18.7 & 22.6 & 72.0 & 61.1 & 59.5 & 60.3 & 28.9 & 29.0 & 14.7 & 19.5 \\
    CADA-VAE \cite{schonfeld2019generalized} & 76.9 & 56.1 & 56.0 & 56.0 & 32.1 & 50.4 & 25.0 & 33.4 & 75.1 & 65.7 & 56.1 & 60.5 & 32.9 & 49.7 & 25.9 & 34.0 \\
    SynSE \cite{gupta2021syntactically} & 71.9 & 51.3 & 47.4 & 49.2 & 31.3 & 44.1 & 22.9 & 30.1 & 68.0 & 65.5 & 45.6 & 53.8 & 29.9 & 61.3 & 24.6 & 35.1 \\
    SMIE \cite{zhou2023zero} & 77.9 & - & - & - & 41.5 & - & - & - & 79.0 & - & - & - & 41.0 & - & - & -\\
    PURLS \cite{zhu2024part} & 79.2 & - & - & - & 41.0 & - & - & - & - & - & - & - & - & - & - & -\\
    SA-DAVE \cite{li2024sadvae} & \textcolor{blue}{\textbf{82.4}} & 62.8 & \textcolor{blue}{\textbf{70.8}} & 66.3 & 41.4 & 50.2 & 36.9 & 42.6 & - & - & - & - & - & - & - & -\\
    STAR \cite{chen2024fine} & 81.4 & \textcolor{blue}{\textbf{69.0}} & 69.9 & \textcolor{blue}{\textbf{69.4}} & \textcolor{blue}{\textbf{45.1}} & \textcolor{red}{\textbf{62.7}} & \textcolor{blue}{\textbf{37.0}} & \textcolor{blue}{\textbf{46.6}} & \textcolor{blue}{\textbf{81.6}} & \textcolor{red}{\textbf{71.9}} & \textcolor{blue}{\textbf{70.3}} & \textcolor{blue}{\textbf{71.1}} & \textcolor{blue}{\textbf{42.5}} & \textcolor{red}{\textbf{66.2}} & \textcolor{blue}{\textbf{37.5}} & \textcolor{blue}{\textbf{47.9}} \\
    \hline
    \textbf{Neuron (Ours)} & \textcolor{red}{\textbf{86.9}} & \textcolor{red}{\textbf{69.1}} & \textcolor{red}{\textbf{73.8}} & \textcolor{red}{\textbf{71.4}} & \textcolor{red}{\textbf{62.7}} & \textcolor{blue}{\textbf{61.6}} & \textcolor{red}{\textbf{56.8}} & \textcolor{red}{\textbf{59.1}} & \textcolor{red}{\textbf{87.8}} & \textcolor{blue}{\textbf{70.6}} & \textcolor{red}{\textbf{75.9}} & \textcolor{red}{\textbf{73.2}} &	\textcolor{red}{\textbf{63.3}} & \textcolor{blue}{\textbf{65.3}} & \textcolor{red}{\textbf{58.1}} & \textcolor{red}{\textbf{61.5}} \\
    \hline
\end{tabular}}
\end{table*}

\begin{table*}
  \caption{State-of-the-art comparisons on NTU RGB+D 120 dataset under the ZSL and GZSL setting. The best and the second-best results are marked in \textcolor{red}{\textbf{Red}} and \textcolor{blue}{\textbf{Blue}}, respectively.}
  \vspace{-10pt}
  \label{table:ntu120}
  \centering
  \resizebox{1.0\textwidth}{!}{
  \begin{tabular}{l|c|ccc|c|ccc|c|ccc|c|ccc}
    \hline
    \multirow{4}{*}{Method} & \multicolumn{8}{c|}{Xusb} & \multicolumn{8}{c}{Xset}\\
    \cline{2-17}
     & \multicolumn{4}{c|}{110/10 Split} & \multicolumn{4}{c|}{96/24 Split} & \multicolumn{4}{c|}{110/10 Split} & \multicolumn{4}{c}{96/24 Split}\\
    \cline{2-17}
    & ZSL & \multicolumn{3}{c|}{GZSL} &  ZSL & \multicolumn{3}{c|}{GZSL} &  ZSL & \multicolumn{3}{c|}{GZSL} &  ZSL & \multicolumn{3}{c}{GZSL} \\
    \cline{2-17}
     & $\mathcal{A}cc$ & $\mathcal{S}$ & $\mathcal{U}$ & $\mathcal{H}$ & $\mathcal{A}cc$ & $\mathcal{S}$ & $\mathcal{U}$ & $\mathcal{H}$ & $\mathcal{A}cc$ & $\mathcal{S}$ & $\mathcal{U}$ & $\mathcal{H}$ & $\mathcal{A}cc$ & $\mathcal{S}$ & $\mathcal{U}$ & $\mathcal{H}$ \\
    \hline
    ReViSE \cite{hubert2017learning} & 19.8 & 0.6 & 14.5 & 1.1 & 8.5 & 3.4 & 1.5 & 2.1 & 30.2 & 4.0 & 23.7 & 6.8 & 13.5 & 2.6 & 3.4 & 2.9\\
    JPoSE \cite{wray2019fine} & 57.3 & 53.6 & 11.6 & 19.1 & 38.1 & 41.0 & 3.8 & 6.9 & 52.8 & 23.6 & 4.4 & 7.4 & 38.5 & 79.3 & 2.6 & 4.9 \\
    CADA-VAE \cite{schonfeld2019generalized} & 52.5 & 50.2 & 43.9 & 46.8 & 38.7 & 48.3 & 27.5 & 35.1 & 52.5 & 46.0 & 44.5 & 45.2 & 38.7 & 47.6 & 26.8 & 34.3 \\
    SynSE \cite{gupta2021syntactically} & 52.4 & 57.3 & 43.2 & 49.5 & 41.9 & 48.1 & 32.9 & 39.1 & 59.3 & 58.9 & 49.2 & 53.6 & 41.4 & 46.8 & 31.8 & 37.9 \\
    SMIE \cite{zhou2023zero} & 61.3 & - & - & - & 42.3 & - & - & - & 57.0 & - & - & - & 42.3 & - & - & -\\
    PURLS \cite{zhu2024part} & \textcolor{red}{\textbf{72.0}} & - & - & - & 52.0 & - & - & - & - & - & - & - & - & - & - & -\\
    SA-DAVE \cite{li2024sadvae} & 68.8 & \textcolor{blue}{\textbf{61.1}} & \textcolor{red}{\textbf{59.8}} & \textcolor{blue}{\textbf{60.4}} & \textcolor{blue}{\textbf{46.1}} & \textcolor{blue}{\textbf{58.8}} &35.8 & \textcolor{blue}{\textbf{44.5}} & - & - & - & - & - & - & - & -\\
    STAR \cite{chen2024fine} & 63.3 & 59.9 & 52.7 & 56.1 & 44.3 & 51.2 & \textcolor{blue}{\textbf{36.9}} & 42.9 & \textcolor{blue}{\textbf{65.3}} & \textcolor{blue}{\textbf{59.3}} & \textcolor{red}{\textbf{59.5}} & \textcolor{blue}{\textbf{59.4}} & \textcolor{blue}{\textbf{44.1}} & \textcolor{blue}{\textbf{53.7}} & \textcolor{blue}{\textbf{34.1}} & \textcolor{blue}{\textbf{41.7}}\\
    \hline
    \textbf{Neuron (Ours)} & \textcolor{blue}{\textbf{71.5}} & \textcolor{red}{\textbf{67.6}} & \textcolor{blue}{\textbf{59.5}} & \textcolor{red}{\textbf{63.3}} & \textcolor{red}{\textbf{57.1}} & \textcolor{red}{\textbf{67.5}} & \textcolor{red}{\textbf{44.4}} & \textcolor{red}{\textbf{53.6}} & \textcolor{red}{\textbf{71.1}} & \textcolor{red}{\textbf{67.5}} & \textcolor{blue}{\textbf{58.9}} & \textcolor{red}{\textbf{62.9}} & \textcolor{red}{\textbf{54.0}} & \textcolor{red}{\textbf{67.0}} & \textcolor{red}{\textbf{44.9}} & \textcolor{red}{\textbf{53.8}}\\
    \hline
\end{tabular}}
\end{table*}

\begin{table*}
  \caption{State-of-the-art comparisons on PKU-MMD I dataset under the ZSL and GZSL setting. The best and the second-best results are marked in \textcolor{red}{\textbf{Red}} and \textcolor{blue}{\textbf{Blue}}, respectively.}
  \vspace{-10pt}
  \label{table:pkummd}
  \centering
  \resizebox{1.0\textwidth}{!}{
  \begin{tabular}{l|c|ccc|c|ccc|c|ccc|c|ccc}
    \hline
    \multirow{5}{*}{Method} & \multicolumn{8}{c|}{Xsub} & \multicolumn{8}{c}{Xview}\\
    \cline{2-17}
     & \multicolumn{4}{c|}{46/5 Split} & \multicolumn{4}{c|}{39/12 Split} & \multicolumn{4}{c|}{46/5 Split} & \multicolumn{4}{c}{39/12 Split}\\
    \cline{2-17}
    & ZSL & \multicolumn{3}{c|}{GZSL} &  ZSL & \multicolumn{3}{c|}{GZSL} &  ZSL & \multicolumn{3}{c|}{GZSL} &  ZSL & \multicolumn{3}{c}{GZSL} \\
    \cline{2-17}
     & $\mathcal{A}cc$ & $\mathcal{S}$ & $\mathcal{U}$ & $\mathcal{H}$ & $\mathcal{A}cc$ & $\mathcal{S}$ & $\mathcal{U}$ & $\mathcal{H}$ & $\mathcal{A}cc$ & $\mathcal{S}$ & $\mathcal{U}$ & $\mathcal{H}$ & $\mathcal{A}cc$ & $\mathcal{S}$ & $\mathcal{U}$ & $\mathcal{H}$ \\
    \hline
    ReViSE \cite{hubert2017learning} & 54.2 & 44.9 & 34.5 & 39.1 & 19.3 & 35.7 & 13.0 & 19.0 & 54.1 & 50.7 & 39.9 & 44.6 & 12.7 & 34.5 & 9.43 & 14.8 \\
    JPoSE \cite{wray2019fine} & 57.4 & 67.0 & 43.0 & 52.4 & 27.0 & 64.8 & 26.5 & 37.6 & 53.1 & 72.9 & 42.5 & 53.7 & 22.8 & 57.6 & 20.2 & 29.9 \\
    CADA-VAE \cite{schonfeld2019generalized} & 73.9 & 76.2 & 51.8 & 61.7 & 33.7 & 69.0 & 29.3 & 41.1 & 74.5 & \textcolor{red}{\textbf{79.9}} & 61.5 & 69.5 & 29.5 & 62.4 & 28.3 & 39.0 \\
    SynSE \cite{gupta2021syntactically} & 69.5 & \textcolor{blue}{\textbf{77.8}} & 40.2 & 53.0 & 36.5 & 71.9 & 30.0 & 42.3 & 71.7 & 69.9 & 51.1 & 59.0 & 25.4 & 61.9 & 22.6 & 33.1 \\
    SMIE \cite{zhou2023zero} & 72.9 & - & - & - & 44.2 & - & - & - & 71.6 & - & - & - & 40.7 & - & - & - \\
    STAR \cite{chen2024fine} & \textcolor{blue}{\textbf{76.3}} & 59.1 & \textcolor{blue}{\textbf{72.3}} & \textcolor{blue}{\textbf{65.0}} & \textcolor{blue}{\textbf{50.2}} & \textcolor{blue}{\textbf{72.7}} & \textcolor{blue}{\textbf{44.7}} & \textcolor{blue}{\textbf{55.4}} & \textcolor{blue}{\textbf{75.4}} & 73.5 & \textcolor{blue}{\textbf{72.2}} & \textcolor{blue}{\textbf{72.8}} & \textcolor{blue}{\textbf{50.5}} & \textcolor{blue}{\textbf{69.8}} & \textcolor{blue}{\textbf{47.5}} & \textcolor{blue}{\textbf{56.5}}\\
    \hline
    \textbf{Neuron (Ours)} & \textcolor{red}{\textbf{89.2}} & \textcolor{red}{\textbf{78.9}} & \textcolor{red}{\textbf{75.9}} & \textcolor{red}{\textbf{77.4}} & \textcolor{red}{\textbf{61.4}} & \textcolor{red}{\textbf{78.3}} & \textcolor{red}{\textbf{56.8}} & \textcolor{red}{\textbf{65.9}} & \textcolor{red}{\textbf{88.2}} & \textcolor{blue}{\textbf{78.7}} & \textcolor{red}{\textbf{77.8}} & \textcolor{red}{\textbf{78.3}} & \textcolor{red}{\textbf{62.2}} & \textcolor{red}{\textbf{75.4}} & \textcolor{red}{\textbf{57.1}} & \textcolor{red}{\textbf{65.0}}\\
    \hline
\end{tabular}}
\end{table*}

\subsection{Evaluation Settings}
\vspace{0.5em}
\noindent\textbf{Protocols.} 
We follow the previous studies \cite{chen2024fine} to conduct the experiments using predefined seen/unseen protocols and tasks in the ZSL and GZSL settings. For NTU 60, we adopt the 55/5 and 48/12 protocols. For NTU 120, they are extended to 110/10 and 96/24 protocols. For PKU-MMD I, we take the 46/5 and 39/12 split strategies.

\vspace{0.5em}
\noindent\textbf{Metrics.}
We evaluate the performance of our method using Top-1 classification accuracy $\mathcal{A}cc=\frac{1}{N} \sum_{i=1}^{N} \mathbb{I}[y_i\in \hat{y}^*]$. In the ZSL setting, we calculate accuracy ($\mathcal{A}cc$) on the $\mathcal{D}_{te}^{u}$. For the GZSL setting, we first compute the seen accuracy ($\mathcal{S}$) on the $\mathcal{D}_{te}^{s}$ and unseen accuracy ($\mathcal{U}$) on the $\mathcal{D}_{te}^{u}$, respectively. Then, the harmonic mean accuracy ($\mathcal{H}$) can be calculated by $\mathcal{H}=(2\times \mathcal{S} \times \mathcal{U})/(\mathcal{S} + \mathcal{U})$.

\subsection{Implementation Details}
We sample the skeleton sequences to 64 frames using the data processing procedure adopted in \cite{chen2024fine}. For the skeleton encoder, we employ the Shift-GCN \cite{cheng2020skeleton}, which is pre-trained on the seen categories. Concurrently, we take the text encoder from the pre-trained \textbf{ViT-L/14@336px} model of CLIP \cite{radford2021learning} to extract semantic embeddings. The optimizer is SGD with a weight decay of 0.0005. The batch size is 64. The base learning rate is 0.1 and reduced with 0.1 multiplied at epochs 10 and 20. The number of $N_s$ and $N_t$ are set to 80. The hyperparameter $\lambda_s$ and $\lambda_t$ are set to 0.0003 and 0.0002. All the experiments are finished using the PyTorch platform on a GeForce RTX 4090 Ti GPU.

\begin{figure}
\begin{center}
    \begin{tabular}{cc}
        \hspace{-12pt}\includegraphics[width=0.55\linewidth]{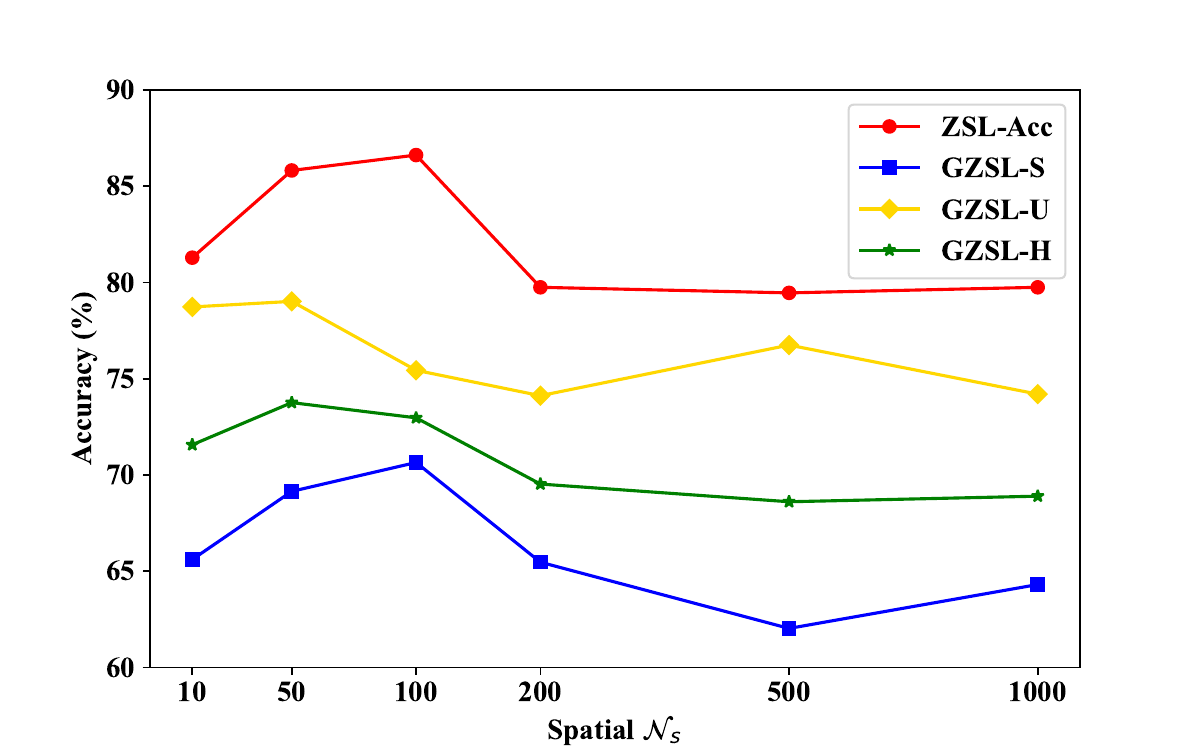} &
        \hspace{-20pt}\includegraphics[width=0.55\linewidth]{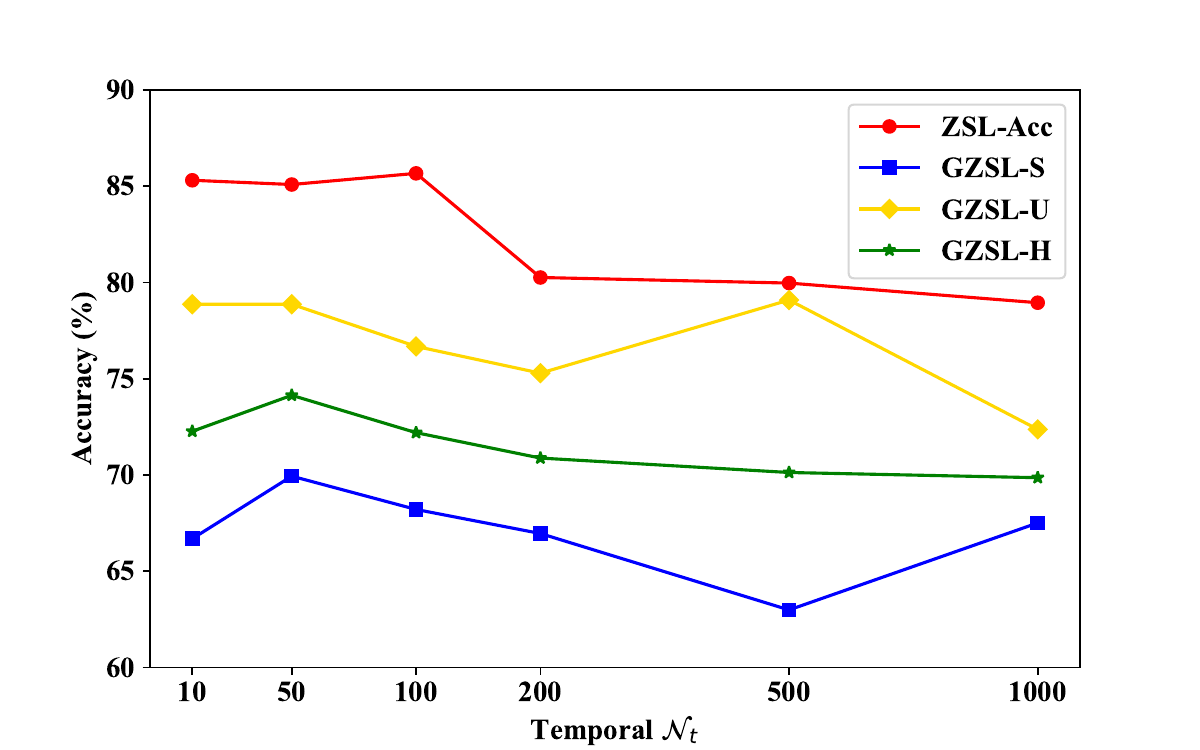}\\
        \hspace{-12pt}\small{(a) Spatial $\mathcal{N}_s$} & 
        \hspace{-20pt}\small{(b) Temporal $\mathcal{N}_t$}\\
    \end{tabular}
\end{center}
\vspace{-15pt}
\caption{The influence of hyper-parameters on the NTU 60.}
\label{fig:hyper-parameters}
\vspace{-10pt}
\end{figure}

\subsection{Comparison with State-of-the-Art}
\vspace{0.5em}
\noindent\textbf{Zero-Shot Learning.}
We evaluate the performance of our method, Neuron, against state-of-the-art approaches in the ZSL setting, as shown in Table \ref{table:ntu60} - \ref{table:pkummd}. Across three distinct datasets representing varied scenarios, our proposed method achieves superior performance. 
Furthermore, the performance of our method is robust across all tasks, demonstrating its ability to learn core spatial-temporal patterns from skeletons that enhance cross-modal correspondences, thereby facilitating better cross-modal transferability.

\vspace{0.5em}
\noindent\textbf{Generalized Zero-Shot Learning.}
As shown in Table \ref{table:ntu60} - \ref{table:pkummd}, our method outperforms previous studies in both seen and unseen categories. 
Additionally, our method achieves an excellent balance between the seen and unseen categories, i.e., mitigating the tendency of the model towards the seen categories (domain bias), improving a considerable margin over others on the value of harmonic mean metrics. Thus, our synergistic framework actually helps to avoid shortcut representation learning, emphasizing spatial structure-related and temporal regularity-dependent feature learning.

\begin{table}[t]
  \caption{Influence of seen-unseen categories settings on NTU RGB+D 60 and PKU-MMD I datasets. The columns of the ZSL and GZSL represent the Top1 accuracy and harmonic mean. The best and the second-best results are marked in \textcolor{red}{\textbf{Red}} and \textcolor{blue}{\textbf{Blue}}.}
  \vspace{-10pt}
  \label{table:known-unknown}
  \centering
  \resizebox{0.45\textwidth}{!}{
  \begin{tabular}{l|c|c|c|c}
    \hline
    \multirow{3}{*}{Method} & \multicolumn{2}{c|}{NTU RGB+D 60} & \multicolumn{2}{c}{PKU-MMD I} \\
     & \multicolumn{2}{c|}{55/5 (Xsub)} & \multicolumn{2}{c}{46/5 (Xsub)} \\
    \cline{2-5}
     & ZSL & GZSL & ZSL & GZSL \\
    \hline
    ReViSE \cite{hubert2017learning} & 54.7 & 27.4 & 48.7 & 32.8 \\
    JPoSE \cite{wray2019fine} & 56.6 & 44.7 & 39.2 & 31.7 \\
    CADA-VAE \cite{schonfeld2019generalized} & 58.0 & 47.1 & 49.0 & 52.7\\
    SynSE \cite{gupta2021syntactically} & 59.9 & 49.9 & 43.5 & 40.4 \\
    SMIE \cite{zhou2023zero} & 64.2 & - & 66.4 & - \\
    STAR \cite{chen2024fine} & \textcolor{blue}{\textbf{77.5}} & \textcolor{blue}{\textbf{62.8}} & \textcolor{blue}{\textbf{70.6}} & \textcolor{blue}{\textbf{67.1}} \\
    \hline
    \textbf{Our Method} & \textcolor{red}{\textbf{84.5}} & \textcolor{red}{\textbf{71.2}} & \textcolor{red}{\textbf{74.4}} & \textcolor{red}{\textbf{69.2}} \\
    \hline
\end{tabular}}
\end{table}

\begin{table}
  \caption{Analysis of different components on NTU RGB+D 60 Xsub dataset under the 55/5 split strategies.}
  \vspace{-10pt}
  \label{tab:analysis_components}
  \centering
  \resizebox{0.48\textwidth}{!}{
  \begin{tabular}{c c c c c c}
    \toprule
    \multicolumn{2}{c}{Spatial Stream} & \multicolumn{2}{c}{Temporal Stream} & \multicolumn{2}{c}{$\mathcal{A}cc$ (\%)} \\
    \cmidrule(r){1-2} \cmidrule(r){3-4} \cmidrule(r){5-6}
    \makecell[c]{Granularity\\Update} & \makecell[c]{Spatial\\Compression} & \makecell[c]{Phase\\Evolving} & \makecell[c]{Temporal\\Memory} & 
    \makecell[c]{ZSL} & \makecell[c]{GZSL}\\ 
    \midrule
    \ding{55} & \ding{55} & \ding{55} & \ding{55} & 79.9 & 65.9 \\
    \checkmark & \ding{55} & \ding{55} & \ding{55} & 81.0 & 66.3 \\
    \checkmark & \ding{55} & \checkmark & \ding{55} & 81.2 & 67.1 \\
    \checkmark & \checkmark & \checkmark & \ding{55} & 81.7 & 69.9 \\
    \checkmark & \checkmark & \checkmark & \checkmark & 86.9 & 71.4 \\
    \bottomrule
    \end{tabular}}
\vspace{-10pt}
\end{table}

\subsection{Ablation Study}
\vspace{0.5em}
\noindent\textbf{Influence of Seen-Unseen Category Settings.}
We evaluate the robustness of our method under the different seen-unseen categories settings. To this end, we re-partition all skeleton categories into three different seen-unseen combinations that are non-overlap, keeping the same as the previous studies \cite{zhou2023zero, chen2024fine}, and then compute the average results of three settings to minimize variance. We report the ZSL and GZSL results in Table \ref{table:known-unknown}. It can be seen that our method consistently achieves state-of-the-art results in different settings, showing excellent stability and robustness. 

\begin{figure*}
\begin{center}
\begin{tabular}{cccc}
    \hspace{-5pt}\includegraphics[width=0.24\linewidth]{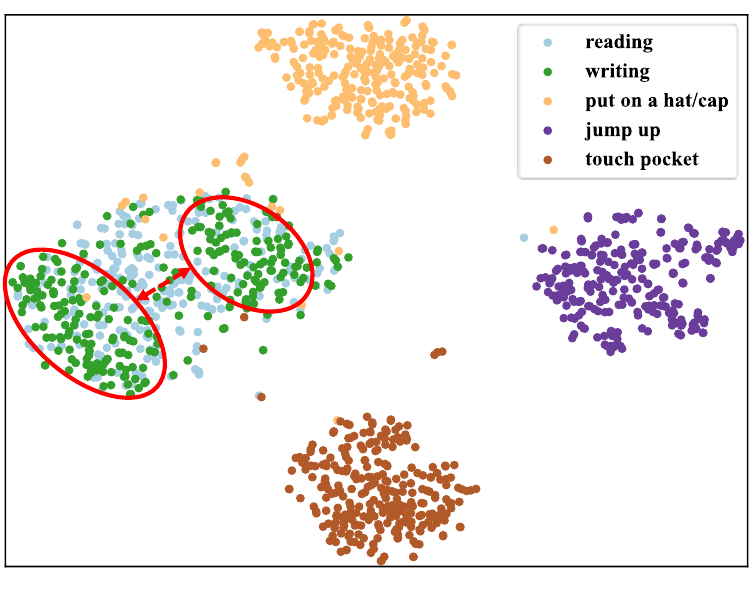} &
    \hspace{-5pt}\includegraphics[width=0.24\linewidth]{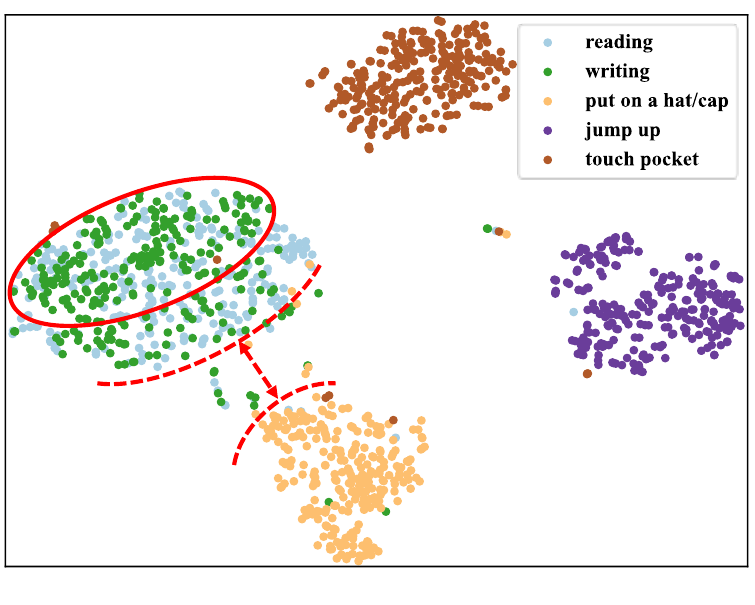} &
    \hspace{-5pt}\includegraphics[width=0.24\linewidth]{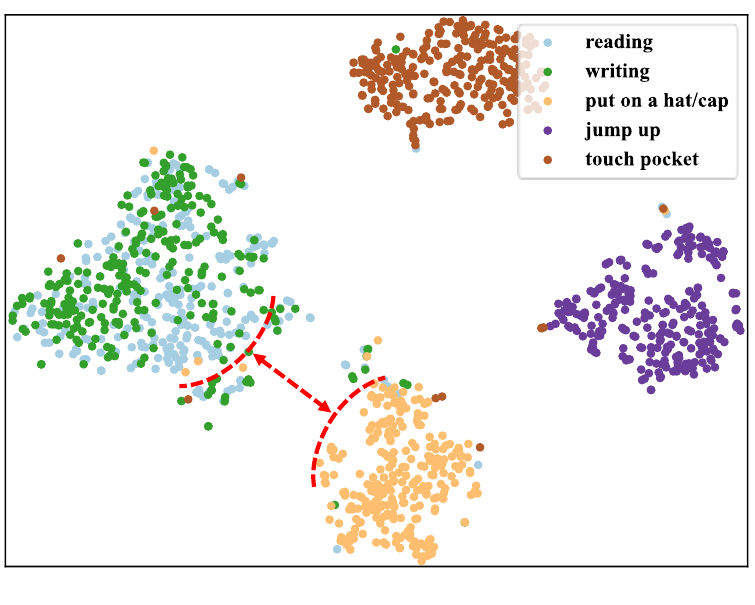} &
    \hspace{-8pt}\includegraphics[width=0.27\linewidth]{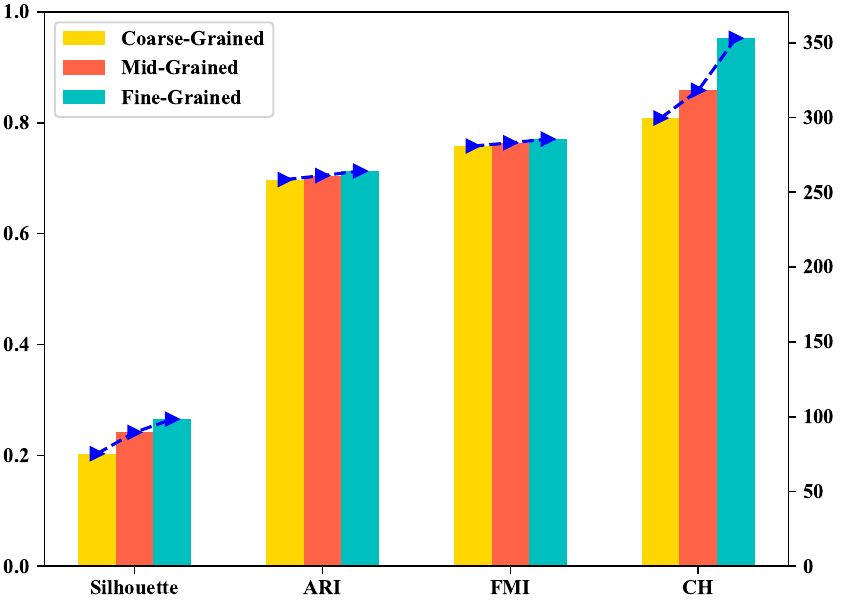}\\
    \hspace{-5pt}\small{(a) Coarse-Grained Phase} & 
    \hspace{-5pt}\small{(b) Mid-Grained Phase} &
    \hspace{-5pt}\small{(c) Fine-Grained Phase} & 
    \hspace{-8pt}\small{(d) Intra-Class Compactness}\\
\end{tabular}
\end{center}
\vspace{-15pt}
\caption{(a) - (c) represent the t-SNE visualization of the skeleton spatial spaces for unseen categories on different phases. The color denotes different unseen categories from the cross-subject task of the NTU 60 dataset under the 55/5 split settings. (d) represents the spatial intra-class compactness metrics of corresponding phases (zoom in for better view).}
\label{fig:compression}
\end{figure*}

\begin{figure*}
\begin{center}
    \includegraphics[width=1.0\linewidth]{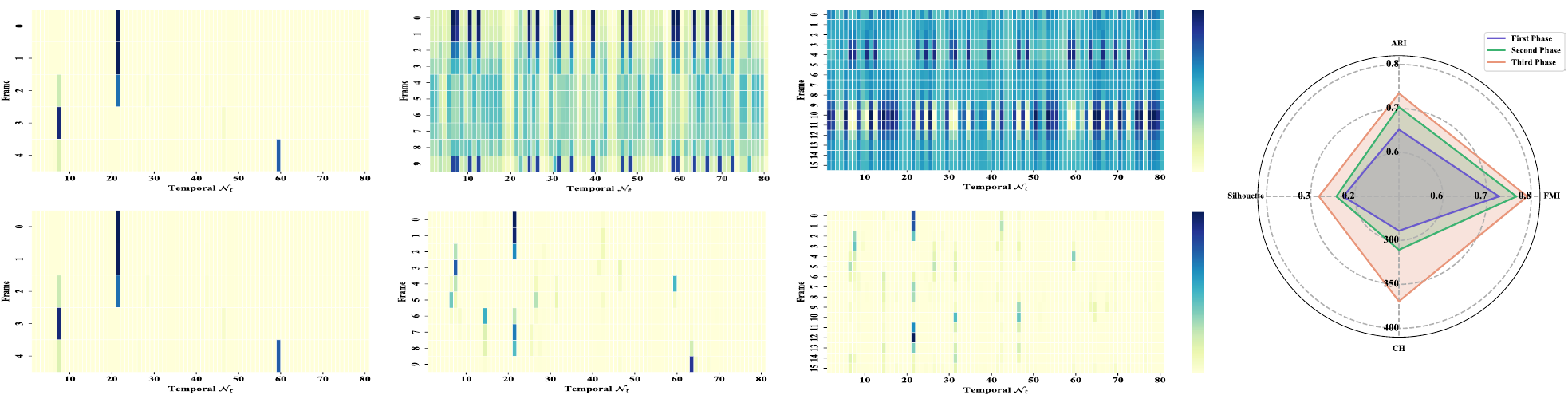}\\ 
    \hspace{35pt}\small{(a) First Phase} \hspace{60pt} \small{(b) Second Phase} \hspace{65pt} \small{(c) Third Phase} \hspace{40pt} \small{(d) Intra-Class Compactness} \\
    \vspace{-5pt}
    \caption{(a) - (c) represent the visualization of updated temporal micro-prototype in each phase for a randomly selected skeleton sequence. The first row denotes the updated process without the proposed temporal memory mechanism, while the second row is equipped. (d) represents the temporal intra-class compactness metrics of corresponding phases (zoom in for better view).}
    \label{fig:memory}
\end{center}
\vspace{-20pt}
\end{figure*}

\vspace{0.5em}
\noindent\textbf{Influence of Components.}
We evaluate the efficiency of different components of our method in Table \ref{tab:analysis_components}. It is noted that the performance of Neuron drops significantly without the temporal memory mechanism, around 5.2\%/1.5\%, showing this technique can mitigate the situation of knowledge oblivion effectively. Additionally, the dual guidance on spatial granularity updating and temporal phase evolving is crucial in progressively controlling the process of cross-modal alignment. Meanwhile, the spatial compression mechanism can improve generalization, increasing the GZSL from 67.1\% to 69.9\%.

\vspace{0.5em}
\noindent\textbf{Influence of $\mathcal{N}_s$ and $\mathcal{N}_t$ in Micro-prototypes.}
As illustrated in Fig. \ref{fig:hyper-parameters} (a) - \ref{fig:hyper-parameters} (b), we observe that the accuracy initially improves and then decreases with the attribute number increases. The best performance of our method can be obtained when the $\mathcal{N}_s$ is around 100 and the $\mathcal{N}_t$ is approximately 50. With further increases in attribute values, accuracy stabilizes and remains consistently high, highlighting the effectiveness of our progressive learning approach.

\subsection{Qualitative Analysis}
\vspace{0.5em}
\noindent\textbf{Visualization of Spatial Compression.}
Here, we plot the t-SNE visualization of unseen skeleton representation distribution at each phase during the spatial compression process, as shown in Fig. \ref{fig:compression} (a) - \ref{fig:compression} (c).
Moving from the coarse-grained to the mid-grained phase, we can see that the compactness of features in "writing" becomes better. Furthermore, the inter-class separability between the "reading" and "put on a hat/cup" actions can also be improved in transitioning from the mid-grained to fine-grained steps. To quantify intra-class compactness, we calculate the compactness of features based on mainstream clustering evaluation metrics, including the silhouette score \cite{rousseeuw1987silhouettes}, Adjusted Rand Index (ARI) \cite{hubert1985comparing}, Fowlkes-Mallows Index (FMI) \cite{fowlkes1983method}, and Calinski-Harabasz Index (CH) \cite{calinski1974dendrite}, as shown in Fig. \ref{fig:compression} (d). 
The intra-class compactness continually improves, underscoring Neuron's effective core feature learning.

\begin{figure}
\begin{center}
    \includegraphics[width=0.8\linewidth]{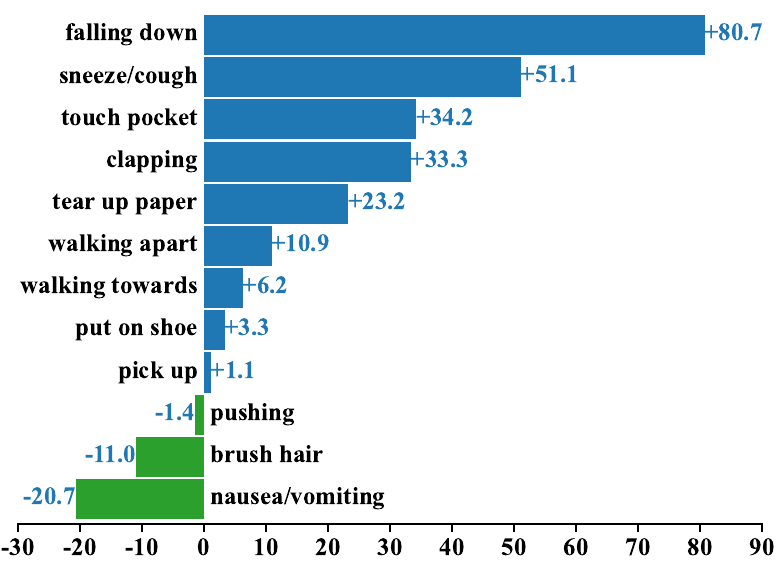}
    \vspace{-10pt}
    \caption{Unseen action classification accuracy compared with STAR \cite{chen2024fine} on the Xsub task of NTU 60 under the 48/12 split setting.}
    \label{fig:compare}
\end{center}
\vspace{-25pt}
\end{figure}

\vspace{0.5em}
\noindent\textbf{Visualization of Temporal Memory.}
As shown in Fig. \ref{fig:memory} (a) - \ref{fig:memory} (c), we draw the updating process of temporal micro-prototypes over multiple steps, comparing results with and without the temporal memory mechanism. In the first row, the updated temporal micro-prototype without the memory mechanism guidance will be chaotic as frames increase, leading to knowledge oblivion. In contrast, the second row exhibits a more effective learning process, where past knowledge is retained while new information is integrated. We also compute the intra-class compactness with the memory mechanism based on the above metrics \cite{rousseeuw1987silhouettes, hubert1985comparing, fowlkes1983method, calinski1974dendrite}, all of them achieve better with the learning process ongoing. 

\vspace{0.5em}
\noindent\textbf{Fine-Grained Action Recognition.}
To compare the performance of our method with the fine-grained method STAR \cite{chen2024fine} more intuitive, we calculate the absolute difference value of recognition accuracy between Neuron and STAR in each unseen category. As shown in Fig. \ref{fig:compare}, we find the accuracy of 9 action categories recognized by our method over the STAR a lot. Specifically, the performance improved for "walking apart" and "walking towards" actions as we paid isolated attention to temporal cues during the training process. Meanwhile, some spatial-oriented actions, e.g., "clapping" and "falling down" actions, are improving significantly. However, some failure cases, such as the "nausea/vomiting" action, are incorrectly classified into the "sneeze/cough" action with highly similar skeletons. Meanwhile, the homogeneous situation still exists between their contextual side information. So, further exploration is still needed for those extremely similar and abstract actions.

\section{Conclusion}
In this paper, we propose \textbf{Neuron}, a novel dy\textbf{\underline{N}}amically \textbf{\underline{E}}volving d\textbf{\underline{U}}al skeleton-semantic syne\textbf{\underline{R}}gistic framework guided by c\textbf{\underline{O}}ntext-aware side informatio\textbf{\underline{N}}, to explore the desirable fine-grained cross-modal correspondence. By introducing the spatial-temporal evolving micro-prototypes to imitate the learning and growth of neurons, the synergistic skeleton-semantic correlations can be established and refined step-by-step under the guidance of contextual side information. In addition, the spatial compression and temporal memory mechanisms are appropriately designed to mitigate the shortcut feature learning and the situation of temporal knowledge oblivion, absorbing the structure-related and regularity-dependent core spatial-temporal representations. Experimental results indicate excellent cross-modal transferability and effective generalizability from seen to unseen categories.
{
    \small
    \bibliographystyle{ieeenat_fullname}
    \bibliography{main}
}


\end{document}